\documentclass{article}
\usepackage{preamble}

\title{Unsupervised learning architecture based on\\ neural Darwinism and Hopfield networks\\ recognizes symbols with high accuracy}
\author{\bf{Mario Stepanik}}
\date{Department of Mathematics, ETH Zürich\\mstepanik@ethz.ch}

\begin{document}

\maketitle

\begin{abstract}
This paper introduces a novel unsupervised learning paradigm inspired by Gerald Edelman's theory of neuronal group selection (``Neural Darwinism''). The presented automaton learns to recognize arbitrary symbols (e.g., letters of an alphabet) when they are presented repeatedly, as they are when children learn to read. On a second hierarchical level, the model creates abstract categories representing the learnt symbols. The fundamental computational unit are simple McCulloch-Pitts neurons arranged into fully-connected groups (Hopfield networks with randomly initialized weights), which are ``selected'', in an evolutionary sense, through symbol presentation. The learning process is fully tractable and easily interpretable for humans, in contrast to most neural network architectures. Computational properties of Hopfield networks enabling pattern recognition are discussed. In simulations, the model achieves high accuracy in learning the letters of the Latin alphabet, presented as binary patterns on a grid. This paper is a proof of concept with no claims to state-of-the-art performance in letter recognition, but hopefully inspires new thinking in bio-inspired machine learning.
\end{abstract}

\section{Introduction}

The late Gerald Edelman, widely recognized for his Nobel-prize-winning research on the chemical structure of antibodies, was known to marvel at the unfathomable individuality and variation of neural structures in human brains (\cite{edelman1998interview}). How could it be that no two brains were alike, in terms of their synaptic circuitry, while they evidently solved similar tasks? Convinced that the Darwinian notions of mutation and selection are fundamental functional principles organizing a variety of biological systems, Edelman introduced a theory of neuronal group selection, later popularized as neural Darwinism, as the unifying algorithm underlying higher brain function (\cite{edelman1982mindful}; \cite{edelman1987neural}). The core of the theory is simple: the fundamental computational unit of the brain is a neural group capable of performing some function. A large number of neuronal groups, strongly coupled internally but weakly among each other, are present initially (e.g., at birth) and are capable of performing a variety of functions. Throughout ontogeny, these groups are selected according to their suitability to solve problems the organism faces. Edelman's previous research had solidified the claim that such a Darwinian process acted as the functional principle underlying our immune system (\cite{rutishauser2014gerald}). Until his death in 2014, his neuroscientific work was dedicated to exploring the idea of a similar selection acting on neuronal groups which are characterized by differences in their connectivity patterns.\newline

While Edelman and colleagues developed a series of automata to formalize the intuition behind neural Darwinism (\cite{edelman1982selective}; \cite{reeke1990synthetic}), no model constructed from building blocks as simple as individual neurons has been proposed, to the best of my knowledge. Instead, previous models have relied on a possible later discovery of a recognition mechanism which explains the functioning of neuronal groups. In this paper, I propose that Hopfield networks, generally appreciated as computational models of associative memory (\cite{hopfield1982neural}), can provide the missing link by exploiting an intuitive, but little discussed, property of their convergence algorithm. In particular, it is demonstrated that the speed of convergence, defined below, can act as a natural and interpretable indicator of ``recognition''. Simulations demonstrate that an automaton combining Hopfield networks, my recognition criterion, and the previous architecture proposed by \textcite{edelman1982selective} can learn to recognize and abstract an arbitrary alphabet solely by repeated presentation of the individual letters with high accuracy. This model may be unique in its ability to integrate the neuronal and functional levels of analysis while retaining complete interpretability, which is of particular interest in machine learning research (\cite{gilpin2018explaining}).\newline

The remainder of this paper is structured as follows: Section \ref{sec:theory} introduces the architecture and the theoretical formulation of the proposed automaton. Section \ref{sec:sim} presents the results of the simulation studies. These findings and the computational and neurobiological properties of the automaton are discussed in Section \ref{sec:dis}. Section \ref{sec:con} concludes.

\section{Theoretical model}
\label{sec:theory}

The automaton presented here consists of three main building blocks: (i) neuronal groups (Hopfield networks), (ii) a recognition repertoire, and (iii) an abstraction repertoire. In this architecture, neuronal groups are the fundamental units responsible for the recognition of inputs. Each neuronal group is randomly initialized (in a sense to be made precise) and a large number of such groups is organized in the loosely coupled recognition repertoire. Intuitively, this repertoire corresponds to the natural diversity (achieved through mutation) among which selection takes place in a Darwinian manner. Upon presentation of a stimulus to all groups, the best-performing one is identified and propagates its recognition ability to nearby groups. On a second hierarchical level, another set of neuronal groups of the same kind (termed the abstraction repertoire) is initialized randomly. These groups do not receive input from external sources (i.e., stimuli), but only from the recognition repertoire to which they are sparsely connected. Over time, these abstraction groups learn to become abstract representations of the presented stimuli. The formal description of these computational properties is presented in the following subsections. The architecture is inspired by previous work by \textcite{reeke1990synthetic}. 

\subsection*{Neuronal groups}

The fundamental computational unit of which groups are composed is a classical McCulloch-Pitts neuron. For such a neuron $k$, let there be $m$ inputs with signals $x_1$ to $x_m$ as well as associated weights $w_{k1}$ to $w_{km}$. The signal (``activity'') of neuron $k$ is obtained by:

\begin{equation} \label{eq:MCP}
    s_k = \varphi\left(\sum_{j=0}^m w_{kj}x_j\right)
\end{equation}

where $\varphi$ is usually a threshold function. We assume this function to be an indicator function such that $s_k=1$ when the sum of weighted inputs is at least $0$ and $s_k=0$ otherwise. These units are arranged in a fully-connected group $i$ with $n$ neurons (a Hopfield network) which is fully characterized by the symmetric $n\times n$ weight matrix $W_i$. The elements $w_{kj}$ of the weight matrix $W_i$ hence correspond to the edge weights in an undirected graph. We constrain self-weights (the diagonal elements of $W_i$) to be $0$ and all other connections, initially, to be either $+1$ (excitatory) or $-1$ (inhibitory). If a neuron is chosen to be updated (as described below), the activity rule in Equation \ref{eq:MCP} is invoked.\newline

It is well known that such a fully-connected network will always converge to a stable state if neurons are updated asynchronously (\cite{hopfield1982neural}; \cite{bruck1990convergence}). That is, if we pick a neuron $k$ at random, sum the weights of all other activated neurons and activate neuron $k$ (i.e., set $s_k=1$) if the sum is greater than zero and deactivate it ($s_k=0$) otherwise, then the network will always converge to a stable $n$-dimensional activity pattern after a finite number of updates. From the study of Ising models in statistical physics, it is known that the set of stable states of a Hopfield network corresponds exactly to the set of local minima in the network's energy function (which is the Hamiltonian of the system\footnote{No external magnetic field/individual thresholds are included in this formulation of the model.}):

\begin{equation} \label{eq:Energy}
    E = -\frac{1}{2}\sum_{k,j}w_{kj}s_ks_j
\end{equation}

It is possible that a given network has multiple local minima, in which case the initial activation pattern determines to which stable pattern the system will converge (``basins of attraction''). In the present model I assume, for simplicity, that neurons are not updated randomly but in an ordered, cyclical manner. This implies that we know precisely when convergence has been reached, namely after a full cycle of $n$ non-updates.\newline

Crucially for this model, the weights in the matrix $W_i$ for each group $i$ are initialized randomly and thus, the stable states of each group are random \emph{a priori}.\footnote{In the software implementation, this corresponds to randomly sampling either +1 or -1 for every connection between neurons (see function \texttt{create\_rand\_weights} in the accompanying code).} Note, however, that the number of update steps to convergence depends heavily on the congruence of the initial activation pattern and the stable activation pattern of the network: if the initial pattern is already a stable state, no updates are required, but the number of updates grows loosely in proportion to the Hamming distance between initial and stable patterns. Hence, the number of updates required for convergence, which we denote $q$, may be interpreted as a measure of ``recognition'': if no updates are required, the network fared well in ``recognizing'' the input pattern. In neurobiological terms, this could correspond to convergence speed, which can be usefully interpreted in the context of the recognition repertoire, presented next.\newline

To build intuition for the connection between a pattern of interest and the initial activation pattern of a neuronal group, consider Figure \ref{fig:Rec}a which depict how a simple $5\times5$ input symbol can be converted to a binary string representing the input activation pattern.

\subsection*{Recognition repertoire}

\begin{figure}[t]
  \centering
  \includegraphics[keepaspectratio, width=\textwidth]{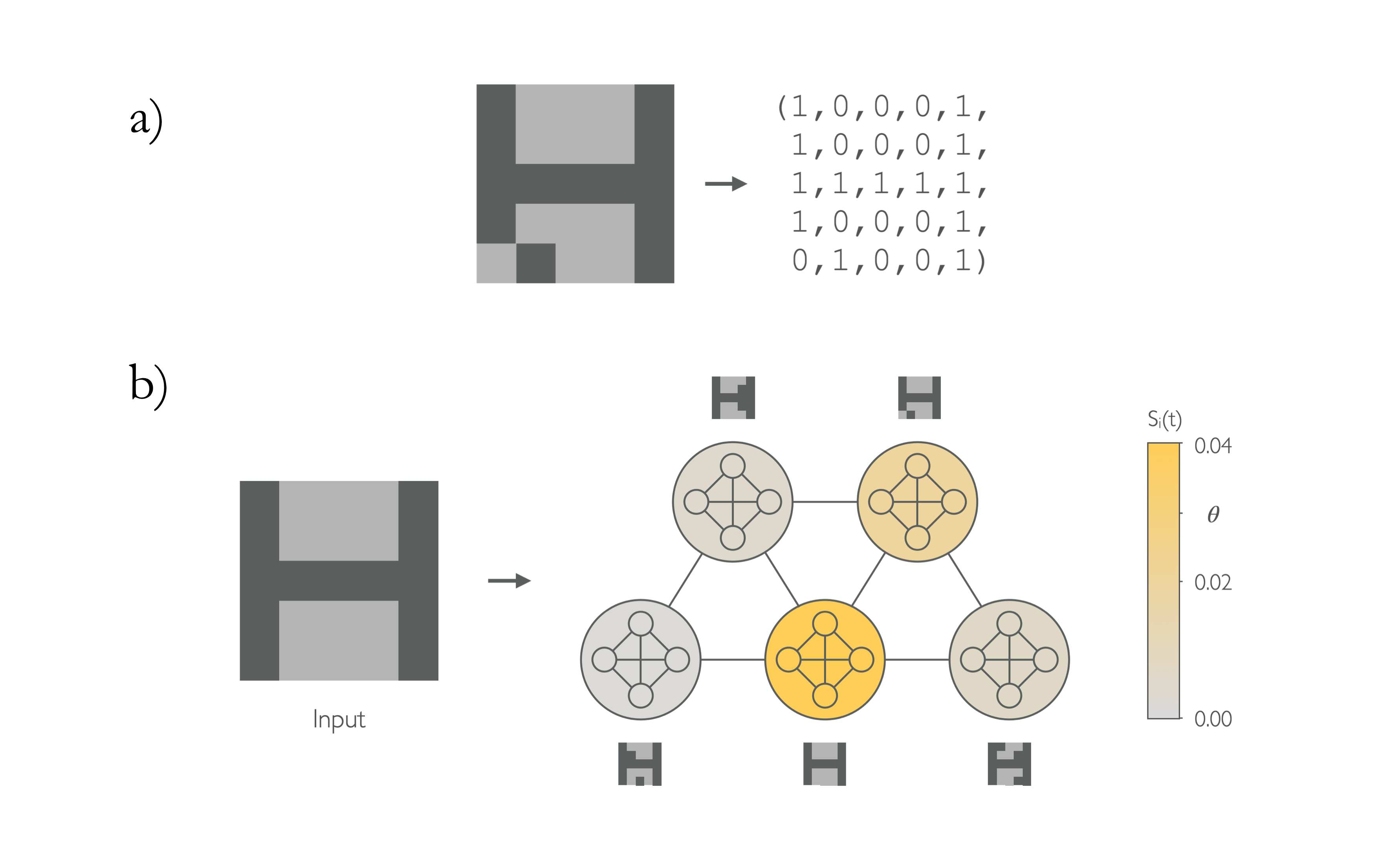}
  \caption{\emph{Top:} A simple $5\times5$ input symbol can be converted to a binary string representing the input activity pattern. \emph{Bottom:} Upon presentation of a symbol (left), a neuronal group $i$ converges in $q_i$ steps, which is translated into an excitation state.}
  \label{fig:Rec}
\end{figure}

Observe that each such neuronal group is intrinsically tuned to ``recognize'' a specific set of symbols, namely those which correspond to binary activation patterns (see Figure \ref{fig:Rec}a) which are local minima of the neuronal group's Hamiltonian. If a pattern identical to a stable state of the network is ``presented'' (that is, the neuronal group is initialized to the binary activation pattern corresponding to this pattern), no updates are required, which we view a ``fast convergence'' in a biological sense.\footnote{Note that this requires the number of neurons in a group, $n$, to be equal to the dimension of the input pattern, which we assume throughout (I return to limitations of this assumption in the discussion section).} We shall now express this idea rigorously and see how it can be exploited in learning.\newline

Let us assume the model evolves in discrete time steps $t$. Between $t$ and $t+1$, an input pattern (stimulus) is presented to a neuronal group $i$, convergence is reached, and the number of updates required for convergence, $q_i$, is recorded. We equip each group with an time-dependent excitation state $S_i(t)$ which inversely depends on $q_i$:

\begin{equation} \label{eq:Excitation}
    S_i(t) = \Phi[q_i(t)] + \omega S_i(t-1) + \varepsilon(t)
\end{equation}

Here, $\Phi[q_i(t)]$ represents a function relating the number of convergence steps at time $t$ to an excitation state. In this case, we take it to be $\frac{1}{q_i}$ if a stimulus is presented at time $t$ and $0$ otherwise. $\omega\in[0,1[$ is an attenuation factor which determines the degree to which previous activation persists across time. $\varepsilon(t)$ is Gaussian noise. Equation \ref{eq:Excitation} essentially translates the performance of a group in recognizing a presented pattern into a scalar value which can be compared to the performance of other groups.\newline

With this formal definition of a neuronal group in hand, we can now define the starting condition of the first hierarchical level, called the recognition repertoire. This repertoire is a collection of $r$ neuronal groups of the kind described above, sparsely connected to each other to allow weight propagation. The intuitive idea behind this setup is illustrated in Figure \ref{fig:Rec}b: suppose we have a large number $r$ of neuronal groups, each initialized randomly to recognize some stimulus in the sense defined above. If a certain set of stimuli is presented repeatedly (e.g., an alphabet), some groups, naturally tuned to recognize the presented symbols, will react by exhibiting an increased excitation state $S$. For example, if the letter ``H'' is presented (as in Figure \ref{fig:Rec}b), a group $i$ which intrinsically possesses an identical convergence pattern will be excited most strongly (illustrated by the neuronal group in the bottom center). If the repertoire of neuronal groups is large enough, there will likely be a group which recognizes any given stimulus.\newline

We can implement a notion of learning by defining a mechanism by which the most excited groups propagate their weight matrix $W$ to a set of nearby groups. This is desirable because a high excitation state indicates the salience of the stimulus which caused the excitation. If a consistent, not too large set of stimuli is presented over time, more neuronal groups will learn to recognize these stimuli over time as their weight matrices become more similar to the weight matrix of the group which initially (through random initialization) recognized the stimulus. Formally, we can describe the weight propagation mechanism as follows:

\begin{equation} \label{eq:UpdateRec}
    W_i(t)=\frac{1}{\max |w_i(t)|}\left[W_i(t-1) + \alpha\sum_l [S_l(t-1)-\theta]_+W_l(t-1)\right]
\end{equation}

where $\alpha$ denotes the learning rate, $\theta$ is a threshold value, the index $l$ enumerates all groups directly connected to $i$ (described in more detail below), and $\max |w_i(t)|$ describes the absolute value of the maximum entry in the new weight matrix, so dividing by it normalizes the resulting matrix to a maximum absolute value of 1 again. The subscript + indicates that the preceding term in brackets is set to 0 if it evaluates to a negative number (to avoid ``negative weight propagation''). Intuitively, each weight matrix $W_i$ is updated by adding a fraction (determined by $\alpha$) of the weight matrices of sufficiently excited (determined by $\theta$) surrounding groups. If a group $j$ is consistently excited (because it recognizes a frequently presented, i.e., salient, stimulus) while group $i$ is not, $j$'s weight matrix will be transferred to group $i$ over time. In the software implementation, I further included a high threshold level of excitation which, if surpassed, prevents a group's weights from being updated to ensure stability.\newline

Following the original concept presented by \textcite{edelman1982mindful}, the neuronal groups are ``topographically'' arranged and connected in the recognition repertoire. This means that during initialization, groups which are determined to be directly connected to each other (and are thus able to receive weight updates) resemble each other in the sense that their weight matrices differ only in a small number of entries. This implies that connected groups recognize similar stimuli. Inspired by the topographical organization in parts of the cortex, \textcite{edelman1982mindful} argued this would facilitate learning. Note that this does not make the initialization less random, as we are simply rearranging the groups to meet this topography criterion: no information about potential stimuli is reflected in the weights in this process. I have adopted this assumption.\newline

In summary, the first hierarchical level of the model operates according to the following steps: first, a recognition repertoire containing $r$ topographically arranged groups of $n$ neurons is created (using the function \texttt{create\_rec\_rep}). Second, an input pattern is presented at each time point $t$, i.e., all activation patterns of the groups in the recognition repertoire are set to the input pattern. All groups converge to an intrinsic stable state and the number of steps to convergence is recorded for each group (function \texttt{present\_pattern\_cum}). Third, at the same time step $t$, groups which converged after a low number of updates are highly excited (Equation \ref{eq:Excitation}) and propagate their weight matrix to less excited groups (Equation \ref{eq:UpdateRec}). Over time, more groups will recognize those stimuli which are presented repeatedly and hence, the recognition repertoire is tuned to the specific set of symbols presented to it without any explicit assumptions about which symbols it might encounter.

\subsection*{Abstraction repertoire}

\begin{figure}[t]
  \centering
  \includegraphics[keepaspectratio, width=\textwidth]{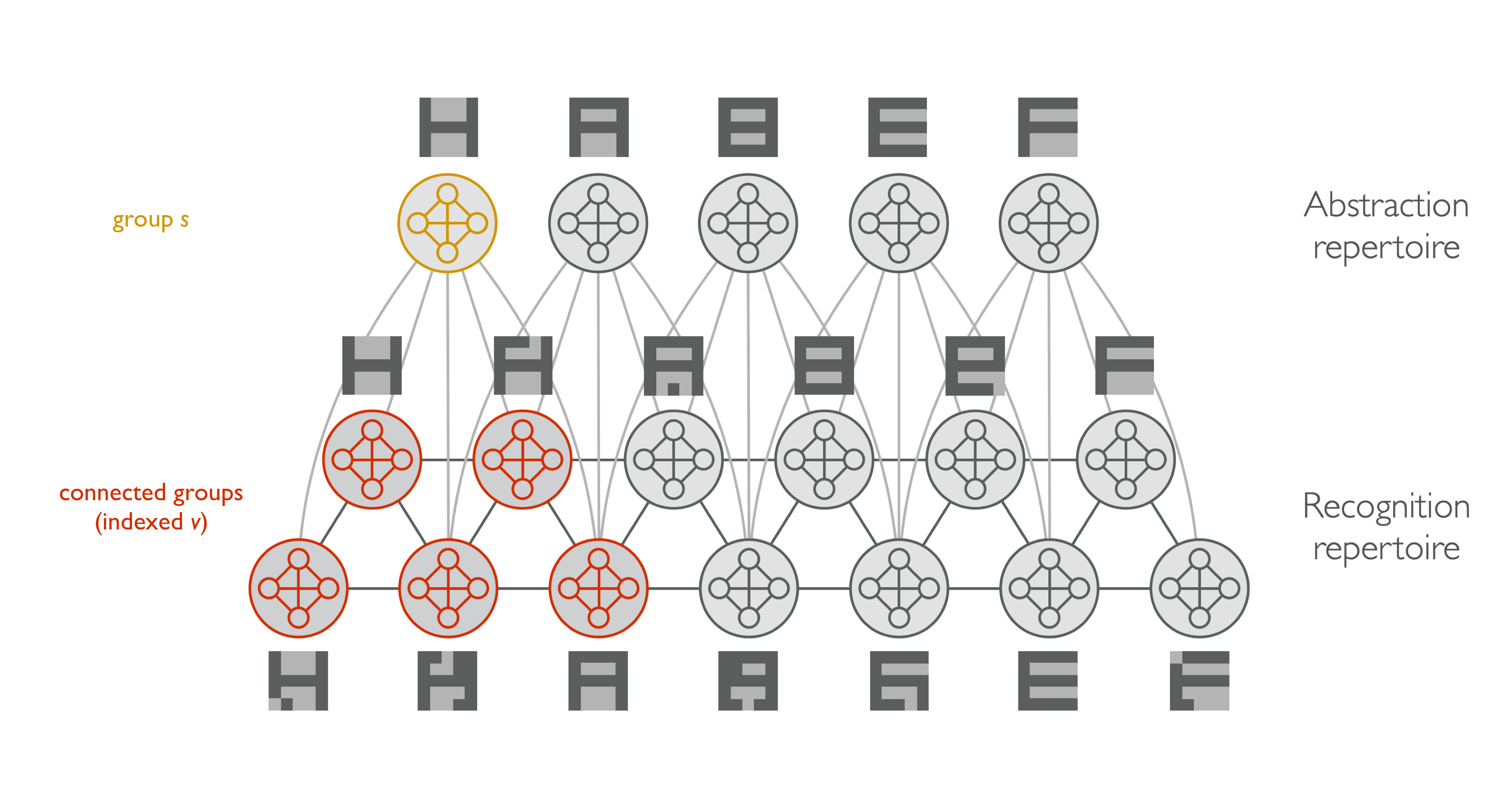}
  \caption{Representation of the network's hierarchical structure. Stimuli are presented to all groups in the recognition repertoire. Sufficient excitation in parts of the recognition repertoire triggers weight propagation to the abstraction repertoire. The connections between an abstraction group $s$ (yellow) and a set of recognition groups, indexed $v$ (red), are shown. The learned patterns that groups are tuned to recognize are schematically depicted near each group.}
  \label{fig:Abs}
\end{figure}

The model is completed by a second hierarchical level termed the abstraction repertoire. This repertoire consists of $a\ll r$ groups of the same kind as those in the recognition repertoire. Each group $s$ in the abstraction repertoire is connected to a small number of (topographically connected) groups in the recognition repertoire (function \texttt{create\_abs\_rep}). These groups do not receive external stimuli directly but only inputs from the recognition repertoire. Each weight matrix of an abstraction group $s$, denoted $W^A_s$, is initialized randomly, as in the recognition repertoire, but is only subject to weight propagation from the groups in the recognition repertoire it is connected to. In contrast, however, an abstraction group is only updated if the cumulative excitation of these connected groups is sufficiently strong (function \texttt{abs\_weight\_updating\_cum}). A schematic representation of the hierarchical structure of the model is depicted in Figure \ref{fig:Abs}.\newline

Intuitively, these groups learn an ``abstract'' category representation of the presented symbols. These are imagined to be our abstract idea of, for example, the letter ``H''. As there are only a small number of abstraction groups, it is unlikely that they are intrinsically tuned to recognize patterns of interest (see simulation results below). Formally, the weight update equation for abstraction groups is given by:

\begin{equation} \label{eq:UpdateAbs}
    W^A_s(t)=\frac{1}{\max |w^A_s(t)|}\left(W^A_s(t-1) + \beta\left[\sum_v [s_v(t-1)W_v(t-1)-\theta]_+-\theta_A\right]_+\right)
\end{equation}

Here, $\beta$ denotes the second-level learning rate, the index $v$ runs over recognition groups connected to the abstraction group $s$, $\theta_A$ is the threshold excitation for weight propagation for the abstraction group, and $\max |w^A_s(t)|$ is again a normalization factor. The subscript + is defined as in Equation \ref{eq:UpdateRec}. This equation expresses the idea that, if the sum of excitations in the connected recognition groups exceeds the threshold $\theta_A$, a fraction (determined by $\beta$) of their weight matrices is added to the abstraction group's weight matrix. If the threshold is not exceeded, no change in $W_s^A$ occurs.\newline

Ideally, these groups are tuned, over time, to the set of symbols presented repeatedly to the recognition repertoire. In the following simulation studies, the accuracy is measured by the number of patterns for which there exists an abstraction groups which learned this pattern, in the sense that the pattern is a stable state of the abstraction group network. This concludes the formal description of the model. In the next section, I will present simulation results.

\section{Simulations}
\label{sec:sim}

The model was implemented using only standard packages in Python. The accompanying code is available \href{https://github.com/ItsaMeMario11/NeuronalGroupAutomaton}{here}.

\subsection{Implementation}

For computational reasons, the simulations were run using patterns of dimension $n=4\times 4=16$ (in contrast to the $5\times5$ patterns depicted in the figures). This size allows for $2^{16}=65,536$ possible patterns. Binary $n$-dimensional vectors for each of the 26 letters of the Latin alphabet were defined (as in Figure \ref{fig:Rec}a).\newline

A topologically organized recognition repertoire is created with a simple stepwise procedure: first, an initial group is created. Then, a number of similar groups (i.e., with a small number of resampled weights) are created and connected to the initial group. A few groups of the newly created ones are sampled as the starting point for the next expansion of the same kind. Some backward connections between two sets of expansions are added. The details of this procedure can be studied in the function \texttt{create\_rec\_rep} and the appropriate auxiliary functions. In the results presented here, a total of 49,961 neuronal groups were created.\newline

The abstraction repertoire was set to consist of 250 neuronal groups. To specify their connections, 250 recognition groups were randomly sampled and each group, its connected groups, and the connected groups of these groups were determined to be the connected groups of a given abstraction group. On average, each abstraction group turned out to be connected to 23 recognition groups. Details can be found in the function \texttt{create\_abs\_rep} and the appropriate auxiliary functions.\newline

The learning procedure is as follows: one of the 26 letters is randomly selected and presented to all neuronal groups in the recognition repertoire. The excitation state of each group is determined using Equation \ref{eq:Excitation} and appropriate weight updates are performed using Equations \ref{eq:UpdateRec} and \ref{eq:UpdateAbs}. For each letter, this process was repeated 10 times to emulate the process of focusing on one letter to learn it (rather than switching swiftly and randomly between them). This procedure is repeated an arbitrary number of times. After each 100 iterations, the number of recognition and abstraction groups which converged to one of the desired letter patterns were recorded to track how many letters were correctly learned. Table \ref{table:params} lists the parameter values, identified through trial and error, which yielded good performance. The thresholds specified as ``see code'' are dynamically generated to identify sensible values: for example, the abstraction threshold depends on the number of recognition groups connected to a given abstraction group, as those with more connected groups would be more easily excitable otherwise. For simplicity, no Gaussian noise is simulated for the excitation state.

\begin{table}[h]
\renewcommand{\arraystretch}{1.1}
\centering
\begin{tabular}{lll}
\textbf{Parameter}  & \textbf{Description} & \textbf{Value} \\ \hline
$\omega$ & Attenuation factor & 0.97 \\
$\alpha$ & Learning rate (recognition) & 0.05 \\
$\theta$ & Propagation threshold (recognition) & see code \\
$\beta$ & Learning rate (abstraction repertoire) & 0.3 \\
$\theta_A$ & Propagation threshold (abstraction) & see code \\
\end{tabular}
\caption{Parameters values and descriptions.}
\label{table:params}
\end{table}

\subsection{Results}

The simulations clearly demonstrate that the model is capable of recognizing, learning, and abstracting symbols in the sense described above. The weight matrices of the neuronal groups in the recognition repertoire adapt over time, leading to a majority of groups converging to a stable state which represents a letter pattern. Furthermore, the groups in the abstraction repertoire were able to learn almost all letters presented.\newline

\begin{figure}[t]
  \centering
  \makebox[\textwidth][c]{\includegraphics[width=1.2\textwidth]{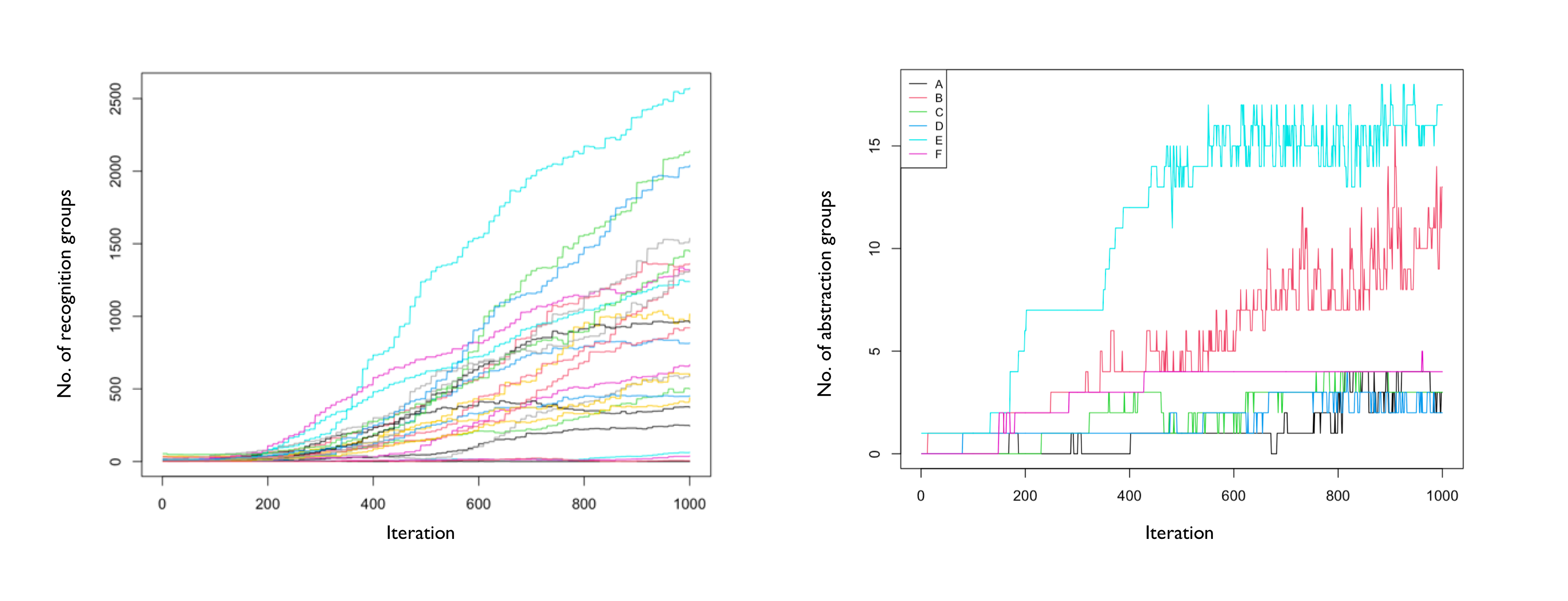}}
  \caption{\emph{Left:} Number of neuronal groups in the recognition repertoire whose stable states are a pattern of interest (letters, as depicted in the legend), out of a total of 49,961. \emph{Right:} Number of abstraction groups whose stable states are a pattern of interest, out of a total of 250. (only first six depicted for visibility). \emph{Note:} The x-axis represents sets of 10 iterations, so the maximum is in fact 10,000.}
  \label{fig:Groups}
\end{figure}

The left panel of Figure \ref{fig:Groups} depicts the evolution of the number of neuronal groups in the recognition repertoire which have one of the letter patterns as their stable state. The lines represent all 26 letters. At time point 0, around 15 $\%$ of letters of the alphabet were not a stable state of any neuronal group and 50 $\%$ of letters were a stable state of a maximum of three groups. Over time, the weight matrices (and thus, the patterns which are stable states of groups) propagate as expected. Note that a small number of letter patterns (here, ``H'', ``Y'', ``Z'') are captured by very few neuronal groups after 10,000 iterations.\newline

\begin{figure}[h]
  \centering
  \makebox[\textwidth][c]{\includegraphics[width=0.6\textwidth]{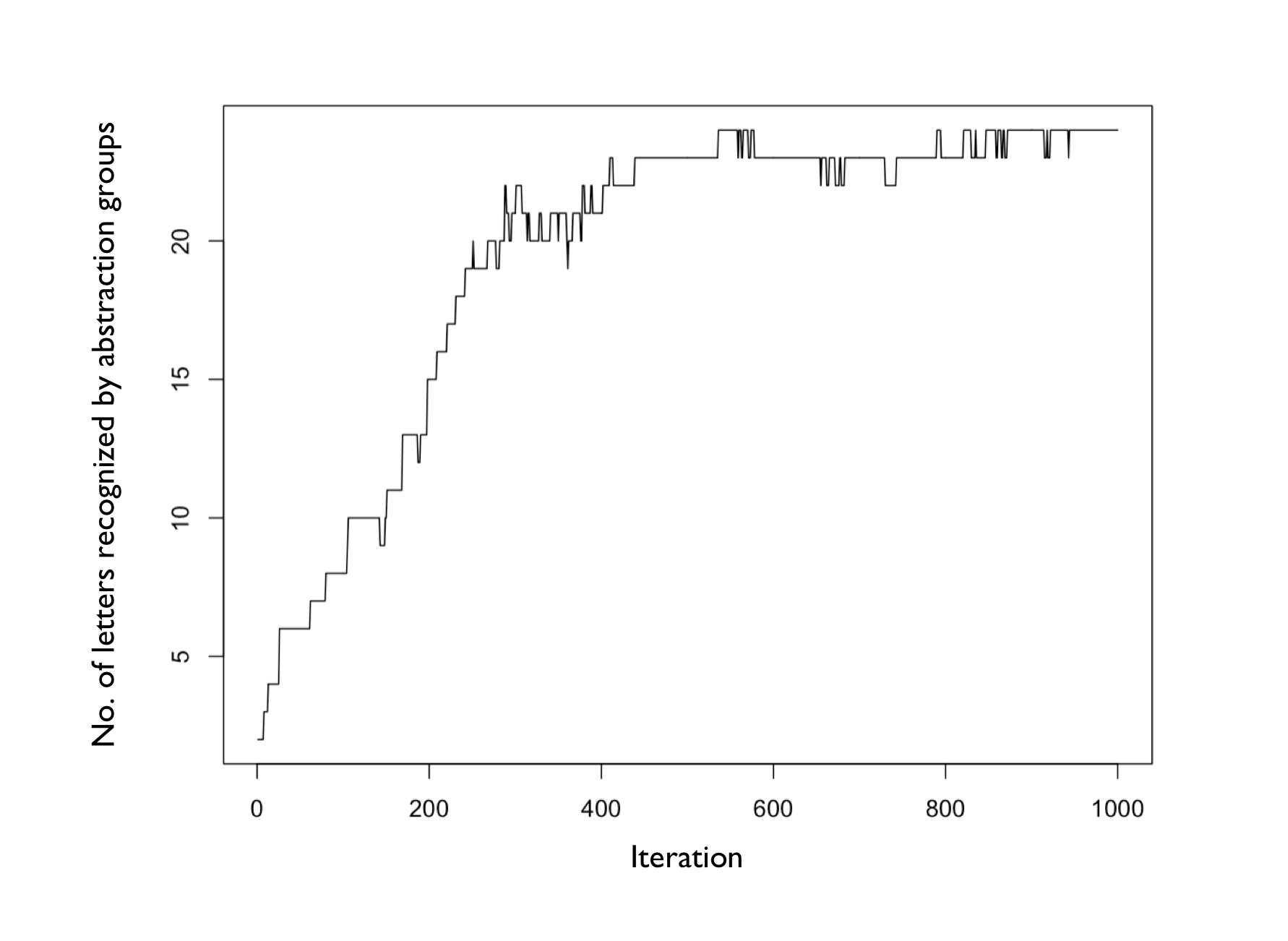}}
  \caption{Number of letter patterns which are stable states of at least one neuronal group in the abstraction repertoire.}
  \label{fig:Letters}
\end{figure}

The right panel of Figure \ref{fig:Groups} depicts the evolution of the number of abstraction groups which have one of the letter patterns as their stable state. For clarity, only the first six letters are presented. At time point 0, only 2 letter patterns are (by chance) stable states of at least one of the 250 abstraction groups (one of them is ``E'', as depicted). After 10,000 iterations, 24 out of 26 letter patterns were stable states of at least one neuronal group in the abstraction repertoire, as depicted in Figure \ref{fig:Letters}. 

\section{Discussion}
\label{sec:dis}

This paper introduced a novel unsupervised learning paradigm inspired by neural Darwinism. The presented automaton learns to recognize arbitrary symbols when they are presented repeatedly and creates abstract category representations for them. I proposed Hopfield networks as the fundamental computational unit implementing the concept by \textcite{edelman1982mindful}.\newline

The model captures a number of key properties of Edelman's original proposal. First, it is a selectionist model: it achieves the task of letter recognition and abstraction by starting from a large variety of randomly initialized fundamental units, among which the stimuli presented to the model ``select'' the most useful ones. No assumptions about the precise patterns to be recognized are made, other than the requirement that they can be represented in binary form in the given dimension. Second, it represents Edelman's observation that selective recognition requires a degenerate (i.e., overlapping) but not redundant repertoire (\cite{edelman1982mindful}). That is, the recognition units need to strike a balance between functional diversity and similarity, which is achieved with a sufficiently large recognition repertoire.\newline

A major appeal of this model is the intuitive interpretation of learning it allows. The system infers which patterns are sufficiently salient that they deserve to be remembered (and abstracted) solely from the frequency in which they are presented. Had we used, for example, letters of the Cyrillic script, these could have similarly been learned. Essentially, it is the regularity with which patterns are presented which matters. Furthermore, we can observe that the reliable recognition of a given pattern comes hand in hand with a loss of flexibility in recognizing new patterns. This is due to the fact that the repeated activation of one group overwrites the recognition ability of another due to weight propagation. This could provide an explanation for the observation that we tend to lose our ability to flexibly acquire new concepts as we age: in a selectionist paradigm, we suffer from a lack of mutation. Finally, the components of the model are easily interpretable for an external observer: the stored patterns of each group can be probed at any time, so the learning mechanism is entirely transparent. This is a major advantage over most machine learning methods, which are frequently seen as black boxes (\cite{rudin2019stop}).\newline

Heuristic considerations also reveal similarities to other theoretical frameworks in neuroscience. Note that this model essentially solves a large number of energy minimization problems as convergence in Hopfield networks is equivalent to the identification of a local minimum in the system's energy landscape. This is reminiscent of the free energy principle, which suggests that agents seek to minimize the free energy (or surprise) through the construction of more accurate models of the environment or actions to modify it (\cite{friston2010free}). In this case, the correct recognition of patterns which occur frequently in the agent's environment serves the purpose of minimizing energy in a concrete manner. Furthermore, the principle of active inference, which states that agents may actively seek out new stimuli in the pursuit of surprise minimization (\cite{friston2016active}), chimes in well with the proposed interpretation of learning: while not simulated explicitly, it is imaginable that accuracy increases even further if we seek out stimuli with the goal of actively learning them when we are struggling to do so. For example, it could be attempted to present those letters which were not yet learned for longer periods of time.\newline

Finally, two properties of neurobiological appeal deserve to be mentioned. First, note that no central coordination of the neuronal groups (e.g., time keeping via a clock) is required: the notion of a ``winning group,'' determined by fastest convergence, is natural and neurobiological implementations are imaginable (although see limitations below). Second, a system as the one presented here is capable of generating reliable output despite relying on (potentially) unreliable computational units. The convergence properties of Hopfield networks hinge primarily on the valence of edge weights (determining whether the connection is excitatory or inhibitory), but not their precise magnitude (\cite{hopfield1982neural}). In contrast to major learning mechanisms prevalent in classical machine learning (e.g., backpropagation), which rely on strong architectural and numerical constraints (\cite{lillicrap2020backpropagation}), the present mechanism lowers the requirements for individual neurons and connections.\newline

Of course, this simple implementation suffers from a number of limitations, both as a computational model and as a theory of brain function. From a computational standpoint, one first has to note that a large number of neurons ($49,961\cdot 16 = 799,376$) are required to fulfil a relatively simple task. Second, the model is currently not robust to alterations of the inputs, such as rotation or distortion. One could argue that a lower hierarchical level could be capable of transforming sensory inputs such that they arrive at the recognition repertoire in the appropriate format. However, this possibility has not been modeled here. From a neurobiological perspective, several questions remain to be answered. First, the precise neural circuits implementing these neuronal groups need to be described. Evidently, axonal projections allow only for unidirectional information flow in real nervous systems, whereas connections in classical Hopfield networks are symmetric. Second, biological systems do not operate in discrete time steps as has been assumed here for modeling purposes. It would be interesting to explore a setting in which updating and weight propagation occur in a dynamical fashion instead. Third, the present account lacks a neurobiological explanation of how convergence speed of a network is translated into a group excitation state, which then allows for the transfer of connection weights to other groups. These questions need to be explored in future work.\newline

\section{Conclusion}
\label{sec:con}

This paper presented an unusual approach to pattern recognition in machine learning. It achieves high accuracy in recognizing and categorizing symbols within a selectionist paradigm, introduced in \textcite{edelman1982mindful}. It represents a proof of concept which can be refined and extended in future work and hopefully inspires researchers to tread new paths in the search for biologically plausible computational models. 

\section*{Accompanying code}
The accompanying code is available \href{https://github.com/ItsaMeMario11/NeuronalGroupAutomaton}{here}.

\section*{Acknowledgments}

I am grateful for the conversations with Kasper Rasmussen, Markus Limbeck, and Anton Schäfer, which improved the quality of this manuscript.

\printbibliography[]

\end{document}